\documentclass[conference]{IEEEtran}
\IEEEoverridecommandlockouts
\usepackage{cite}
\usepackage{amsmath,amssymb,amsfonts}
\usepackage{algorithm}
\usepackage{algpseudocode}
\usepackage{graphicx}
\usepackage{xcolor}
\usepackage{listings}
\usepackage{float}
\floatstyle{plaintop}
\newfloat{codelisting}{tbp}{lol}
\floatname{codelisting}{Listing}
\usepackage{booktabs}
\usepackage{url}
\usepackage[hidelinks]{hyperref}
\usepackage{fancyhdr}
\fancypagestyle{arxivnotice}{%
  \fancyhf{}%
  \fancyfoot[C]{\parbox{\textwidth}{\centering\footnotesize
    This work has been accepted for publication in AIxSET 2026. Copyright may be
    transferred without notice, after which this version may no longer be accessible.}}%
}

\lstdefinestyle{par}{basicstyle=\footnotesize\ttfamily,breaklines=true,
  columns=fullflexible,keepspaces=true,frame=single,framesep=3pt,
  xleftmargin=4pt,xrightmargin=2pt,aboveskip=10pt,belowskip=6pt,
  captionpos=t,abovecaptionskip=4pt,belowcaptionskip=14pt}
\begin{document}

\title{Planner-as-Router: Joint Plan-Time Model Routing\\
for Cost-Efficient Multi-Agent Workflows}

\author{
\IEEEauthorblockN{Vivek Kumar Singh}
\IEEEauthorblockA{\textit{Independent Researcher}\\ McKinney, TX, USA\\
0009-0002-9350-3017\\ vivekksingh.nov12@gmail.com}
\and
\IEEEauthorblockN{Preeti Priyam}
\IEEEauthorblockA{\textit{Independent Researcher}\\ McKinney, TX, USA\\
0009-0000-9423-741X\\ preetipriyam12@gmail.com}
\and
\IEEEauthorblockN{Gautam Bhowmick}
\IEEEauthorblockA{\textit{Independent Researcher}\\ Chicago, IL, USA\\
0009-0001-9424-7826\\ bhowmickgautam@gmail.com}
}
\maketitle
\thispagestyle{arxivnotice}

\begin{abstract}
Running large language model (LLM) agents in production gets expensive fast.
A frontier model (the largest, most capable tier) is accurate but can cost
25 times what a small model costs per token, and the gap compounds once a
workflow chains several calls together. Planner-as-Router (PaR) attacks this
from a different angle. Instead of leaving model-tier selection to some
component downstream, it folds the choice into planning itself. As the planner
breaks a query into subtasks, it also assigns each one a model size
tier---small, mid, or frontier, ordered by capability and price---so the
dependencies between subtasks are visible before any specialist runs. Unlike
per-call routers such as cascade routing, which look at one node at a time,
PaR sees the whole workflow up front and needs no separate router model or
training data. We evaluate PaR with EntBench, a benchmark of 54 enterprise
agentic tasks across seven classes, graded by actually running the generated
Structured Query Language (SQL) and MongoDB queries against live databases.
Over 1,157 evaluations spanning eight routers and three seeds, PaR stays on the
observed cost--accuracy frontier. It matches a sink-frontier heuristic (frontier
model on terminal nodes only) in accuracy at comparable cost and a faithful
FrugalGPT cascade at lower cost, and cuts cost 44\% against all-frontier routing
while giving up 2.9 points of accuracy. Several accuracy gaps fall inside the
plus-or-minus six-point confidence interval of a 54-task study, so we frame
PaR's advantage as frontier position rather than a clean accuracy win. We also
report a preliminary observation, not a validated result: a small pilot hints
that cheap routing may carry a hidden compounding penalty on compositional
workflows, which we frame as a hypothesis for future measurement. PaR, EntBench,
and all evaluation code are open source.
\end{abstract}

\begin{IEEEkeywords}
Agentic workflows, benchmarking, cost optimization, enterprise AI, large
language models, model routing, multi-agent systems, workflow orchestration
\end{IEEEkeywords}

\section{Introduction}
Running LLM agents in production gets expensive in a hurry. A frontier model
like Claude Opus runs 5 to 25 times the per-token cost of a small model like
Claude Haiku. That is easy to absorb on one call, but real agentic workflows
rarely make one call. They decompose a query into several specialist steps run
in sequence, and five frontier calls can run an order of magnitude past the
equivalent small-model pipeline. Across a production deployment, always reaching
for the strongest model stops being a rounding error and becomes the dominant
line item. The question is how to budget that well: which steps need a frontier
model, and which can be served far more cheaply without dragging down the answer.

Most routing methods work one call at a time. FrugalGPT~\cite{frugalgpt} is the
classic example: start cheap, and escalate to a stronger model only when
confidence drops. On its own that structure is sensible. Workflows complicate it,
because the output of one node feeds the next. A small model that returns a
slightly wrong answer at step one hands that error to everything downstream,
however capable the later models are, and a per-call router never sees it coming
because it only ever looks at the node in front of it.

Three families define the current state of the art, and none assigns tiers
jointly across a dependency-coupled workflow before execution. Per-call cascades
such as FrugalGPT~\cite{frugalgpt} are the deployed default: cheap-first,
escalate on low confidence, one call at a time. Learned query-level routers such
as RouteLLM~\cite{routellm} and Hybrid-LLM~\cite{hybridllm} report the strongest
published accuracy--cost trade-offs but require labeled training data and route a
whole query to one model. Agent-level routers such as MasRouter~\cite{masrouter}
are the closest prior art, but assign one model per agent rather than per
subtask. Each decides in isolation from the workflow structure that plan time
makes visible.

PaR closes that gap by moving routing into planning. The planner does not hand
tier selection to a later component; it picks the tier for each subtask as it
writes the plan. Having the whole query in view lets it reason about structure,
so a subtask whose result feeds several others (high fan-in) can get a stronger
tier while a simple leaf retrieval stays cheap. PaR needs no trained router and
no extra data, since everything lives in a planner prompt plus a tier-aware
dispatcher, and it drops into graph-based frameworks with little friction.

This paper makes two contributions. First, we introduce PaR, a training-free
pattern that moves tier assignment into the planning phase, with an open-source
implementation; we evaluate it and show it sits on the observed cost--accuracy
frontier, matching a sink-frontier heuristic at comparable cost and a faithful
FrugalGPT cascade at lower cost, and cutting cost 44\% against all-frontier
routing for a 2.9-point accuracy drop. Second, we release EntBench, whose
execution-based grading supports per-node measurements most benchmarks cannot.
Separately, we report a preliminary compounding-error signal on compositional
workflows as an observation that motivates future measurement, not as a
validated result (Section~\ref{sec:compounding}).

\section{Related Work}\label{sec:related}
The cascade idea comes from FrugalGPT~\cite{frugalgpt}, which saved money on
single-call natural language processing (NLP) benchmarks by trying cheap models
first and escalating on low confidence. We reimplement it faithfully with a
lightweight scorer. A second line routes a whole query to one model with a
learned policy: RouteLLM~\cite{routellm} trains on preference data,
Hybrid-LLM~\cite{hybridllm} learns a query-level difficulty router, and
RouterBench~\cite{routerbench} standardizes their evaluation.
AutoMix~\cite{automix} is closest to our scorer, escalating when a small model's
self-verification fails. All share PaR's cost--accuracy goal but differ in
granularity: they send a whole query to one model with a trained classifier,
while PaR assigns a tier per subtask through prompted reasoning and trains
nothing. MasRouter~\cite{masrouter} is closest in spirit but uses a learned
policy and routes per agent. Sindhu and Arumugam~\cite{sindhu} propose the
cost-aware utility $U = A/C^{\alpha}$ we adopt, though for single-step rather
than multi-step selection.

Agent benchmarks usually report end-to-end success and bury the routing decision
inside it. AgentBench~\cite{agentbench} spans web, coding, and database tasks but
evaluates neither per-node tier routing nor cost. BIRD~\cite{bird} and
Spider~\cite{spider} use execution-based grading for text-to-SQL, which shaped
our evaluator, but both are single-step with no routing. $\tau$-bench~\cite{taubench}
and GAIA~\cite{gaia} measure tool-use and general-assistant success but neither
runs structured queries against live databases or breaks down cost. EntBench
fills that gap. AutoGen~\cite{autogen} and LangGraph~\cite{langgraph} support
multi-agent workflows but leave model selection hardcoded per agent.
PaR is designed to compose with such graph-based frameworks: because it is a
planner prompt plus a dispatcher over a typed subtask DAG, its tier assignment
can be layered onto a framework's existing node/edge execution without
retraining. Our reference implementation realizes this as a standalone
topological dispatcher over the planner's DAG; a framework-native integration is
straightforward and left to the released code.
Production agent systems may already perform some form of plan-time model
selection without a published description; we therefore scope our novelty claim
to the published routing literature, where per-subtask tier assignment
conditioned on the full plan has not, to our knowledge, been formalized or
evaluated.

\section{Planner-as-Router Architecture}\label{sec:arch}
\begin{figure*}[t]
\centering
\includegraphics[width=0.82\textwidth]{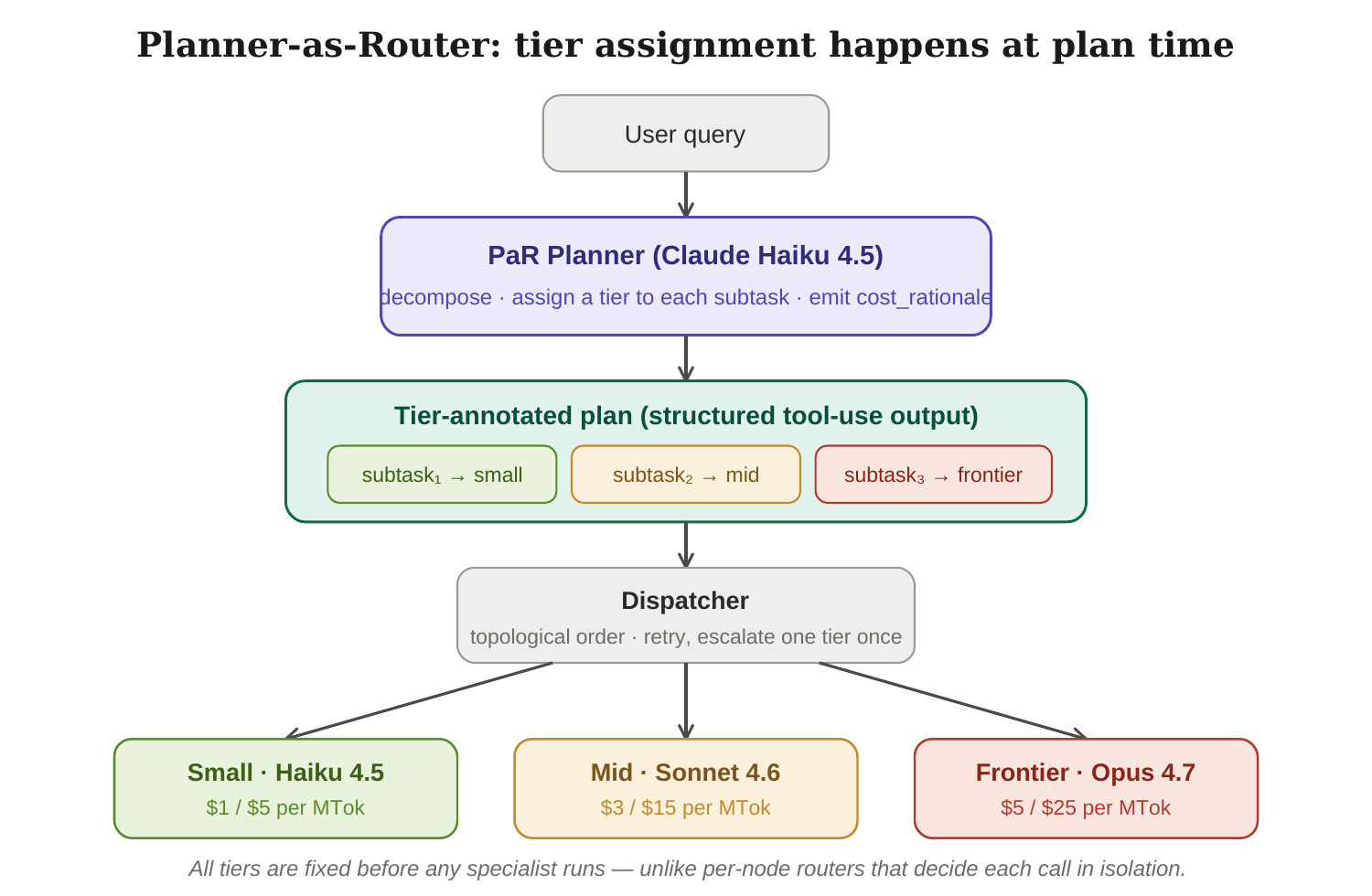}
\caption{Planner-as-Router: tier assignment happens at plan time. The planner
assigns a model tier to every subtask before any specialist runs; the dispatcher
executes subtasks in dependency order, escalating one tier once on failure.}
\label{fig:arch}
\end{figure*}

Consider an agentic system that answers a query $q$ by running a workflow of
specialist agents, each performing one operation: generating SQL, building a
MongoDB pipeline, extracting fields, or reconciling two sources. Each call is
served by a model from tiers $T = \{\text{small}, \text{mid}, \text{frontier}\}$,
ordered by capability and price. The workflow is a directed acyclic graph (DAG):
each node is a subtask with a specialist type and an assigned tier, and an edge
$u \rightarrow v$ means $v$ consumes $u$'s output. Fig.~\ref{fig:arch} shows the
plan-time flow.

The routing problem is to choose the tier for every node to maximize the
cost-aware utility
\begin{equation}
U \;=\; \frac{A}{C^{\alpha}},
\end{equation}
where $A$ is end-to-end accuracy, $C$ total cost, and $\alpha$ the cost
sensitivity. The difficulty is that $A$ is not separable across nodes: a
downstream node's accuracy depends on the tiers chosen upstream. Single-node
routers decide each node in isolation; PaR instead assigns tiers jointly,
conditioning every decision on the full workflow structure and query context,
both available at plan time.

Formally, tier assignment is a constrained optimization over the plan DAG: choose
a tier $t_v \in T$ for each node $v$ to maximize expected end-to-end accuracy $A$
subject to a cost budget, where node accuracies are \emph{dependent}---a
downstream node's success is conditioned on the tiers chosen upstream. The
single-step utility $U = A/C^{\alpha}$ of Sindhu and Arumugam~\cite{sindhu}
scores a completed assignment but does not by itself prescribe one over a
dependency graph, because $A$ is not separable across nodes. Solving the
assignment exactly would require the joint success distribution over the DAG,
which is unavailable at plan time. PaR approximates the solution in a single
planner forward pass: the planner conditions on the whole decomposition at once
and emits all tiers jointly, rather than optimizing each node in isolation as
per-call and query-level routers do. We do not claim optimality; we claim that a
holistic single-pass assignment empirically lands on the cost--accuracy frontier
(Section~\ref{sec:results}).

\subsection{Planner: Prompt Contract and Structured Output}
The planner runs on Claude Haiku~4.5 (small tier,
\texttt{claude-haiku-4-5-20251001}) and is held fixed across every router, so any
accuracy or cost difference reflects tier assignment alone rather than planner
capability. It is invoked with a forced structured-output tool call
(\texttt{tool\_choice} pinned to a single \texttt{produce\_plan} tool,
\texttt{max\_tokens}=1000), so the model cannot return prose in place of a plan.
The system prompt is marked for ephemeral prompt caching, since it is identical
across all tasks.

The tool schema is the plan contract. Every subtask carries five fields: a unique
\texttt{id}; a natural-language \texttt{description}; a \texttt{specialist} from
the closed set \{\texttt{sql\_gen}, \texttt{mongo\_query}, \texttt{extract},
\texttt{cross\_recon}, \texttt{multitool\_plan}, \texttt{policy\_action}\}; a
\texttt{tier} from \{\texttt{small}, \texttt{mid}, \texttt{frontier}\}; and a
\texttt{depends\_on} list of upstream ids. The plan additionally carries a single
\texttt{cost\_rationale} string, marked \emph{required}, that forces the planner
to justify both the subtask count and every tier choice in the same call. Tiers
map to a fixed model and price: small to Haiku~4.5 (USD 1/5 per million
input/output tokens), mid to Sonnet~4.6 (USD 3/15), frontier to Opus~4.7 (USD
5/25).

\begin{codelisting}[t]
\caption{Planner \texttt{produce\_plan} tool schema (abbreviated).}
\label{lst:schema}
\vspace{10pt}
\begin{lstlisting}
produce_plan(
  subtasks: [ {
    id: string, description: string,
    specialist: enum{sql_gen, mongo_query, extract,
                cross_recon, multitool_plan, policy_action},
    tier: enum{small, mid, frontier},
    depends_on: [string]
  } ],
  cost_rationale: string   // required
)
\end{lstlisting}
\end{codelisting}

Listing~\ref{lst:plan} shows a real plan for a Cross-Recon task reconciling
licensed-seat counts in a MongoDB contracts collection against actual assignments
in a SQL table. The two independent retrieval subtasks are assigned the small
tier; the high-fan-in reconciliation that consumes both is lifted to mid, and the
\texttt{cost\_rationale} field records why.

\begin{codelisting}[t]
\caption{A real PaR plan for a Cross-Recon task.}
\label{lst:plan}
\vspace{10pt}
\begin{lstlisting}
{ "subtasks": [
  {"id":"s1","specialist":"mongo_query","tier":"small",
   "depends_on":[],"description":"contracted seats"},
  {"id":"s2","specialist":"sql_gen","tier":"small",
   "depends_on":[],"description":"count assigned seats"},
  {"id":"s3","specialist":"cross_recon","tier":"mid",
   "depends_on":["s1","s2"],"description":"reconcile"} ],
  "cost_rationale":"s1/s2: single-source reads -> small;
   s3: fuses both, decides -> mid." }
\end{lstlisting}
\end{codelisting}

\subsection{Complexity Classifier}
To avoid paying for a full planning call on trivial queries, PaR first runs a
lightweight Haiku classifier (cached system prompt, \texttt{max\_tokens}=10) that
labels the query \textsc{simple} or \textsc{complex}. A \textsc{simple} label
short-circuits planning: PaR emits a single-subtask plan pinned to the small
tier, with the specialist inferred from the task class, and never invokes the
full planner. \textsc{complex} queries fall through to the full planner of
Section~III-A. This split matters for interpreting cost: part of PaR's saving on
single-step classes comes from the classifier short-circuit and part from the
planner's tier reasoning on multi-step classes. The ablation in
Section~\ref{sec:ablation} isolates the latter by removing the cost-rationale
field while leaving the classifier untouched.

\subsection{Dispatcher and Escalation}
The dispatcher (Algorithm~\ref{alg:dispatch}) executes the plan in dependency
order. At each step it selects the ready set---subtasks whose \texttt{depends\_on}
ids are all satisfied---and runs each at its planner-assigned tier, passing
upstream node outputs as context. Escalation follows a single fixed policy
applied identically to every router, so retry effort is a control rather than a
confound: a failing subtask is retried at its assigned tier up to three times,
then escalated exactly one tier and retried once. If escalation also fails, the
node is recorded null and the null propagates to dependents. A running cost
ceiling (a USD~75 kill switch) bounds worst-case spend per task. Cost is
accumulated per node from measured input, output, and cached token counts at the
node's tier price; the planner call is charged at the small-tier price. Reported
per-task cost is specialist-only and excludes the planner call, applied as an
identical additive offset across all routers so that Pareto positions and deltas
are unaffected.

\begin{algorithm}[t]
\caption{PaR Dispatcher}\label{alg:dispatch}
\begin{algorithmic}[1]
\Require Plan $P$ (subtasks, dependencies, assigned tiers); max same-tier retries $R$
\State topologically sort the subtasks of $P$ into $\langle v_1,\dots,v_k\rangle$
\For{each subtask $v_i$ in dependency order}
  \If{any dependency of $v_i$ returned \textbf{null}}
    \State record $v_i \leftarrow$ \textbf{null} and continue to the next subtask
  \Else
    \State $t \leftarrow$ assigned tier of $v_i$; \; $r \leftarrow 0$
    \State $out \leftarrow \textsc{Run}(v_i, t)$
    \While{$out$ failed \textbf{and} $r < R$}
      \State $out \leftarrow \textsc{Run}(v_i, t)$; \; $r \leftarrow r+1$
    \EndWhile
    \If{$out$ failed \textbf{and} $t \neq$ frontier}
      \State $t' \leftarrow \textsc{Escalate}(t)$; \; $out \leftarrow \textsc{Run}(v_i, t')$
    \EndIf
    \State record $out$ (\textbf{null} if still failed)
  \EndIf
\EndFor
\end{algorithmic}
\end{algorithm}

Algorithm~\ref{alg:dispatch} reads as follows. Line~1 topologically sorts the
plan so that every node runs only after its dependencies; independent subtasks
share a level and could be dispatched concurrently, though our implementation
runs them sequentially. Lines~3--5 propagate failure: a node whose dependency
returned null is itself recorded null without being run, which keeps a single
upstream failure from being charged as multiple downstream attempts and isolates
the routing signal from cascade noise---the same isolation the compounding
analysis of Section~\ref{sec:compounding} later measures. Lines~6--10 are the
fixed retry control: the node runs at its assigned tier and, on failure, retries
there up to $R$ times before any escalation, so no router can look artificially
strong by retrying harder. Lines~11--13 permit exactly one one-tier escalation.
We cap escalation at one tier deliberately---unbounded escalation would let every
router converge on the frontier model and erase the comparison we are trying to
make. Line~14 records the node null if escalation also fails; a null costs the
router the node's accuracy without further spend, which is why frontier-heavy
routers do not dominate on cost even when they are more robust.

\section{EntBench}\label{sec:entbench}
EntBench has 54 tasks across seven classes, split into capability tasks that
probe a single tier and compositional tasks that chain several specialists.
Table~\ref{tab:classes} lists the classes. The design supports per-node
measurement: because tasks decompose into typed subtasks graded by execution, we
can attribute success or failure to specific tier assignments rather than an
opaque end-to-end score.

\begin{figure*}[t]
\centering
\includegraphics[width=\textwidth]{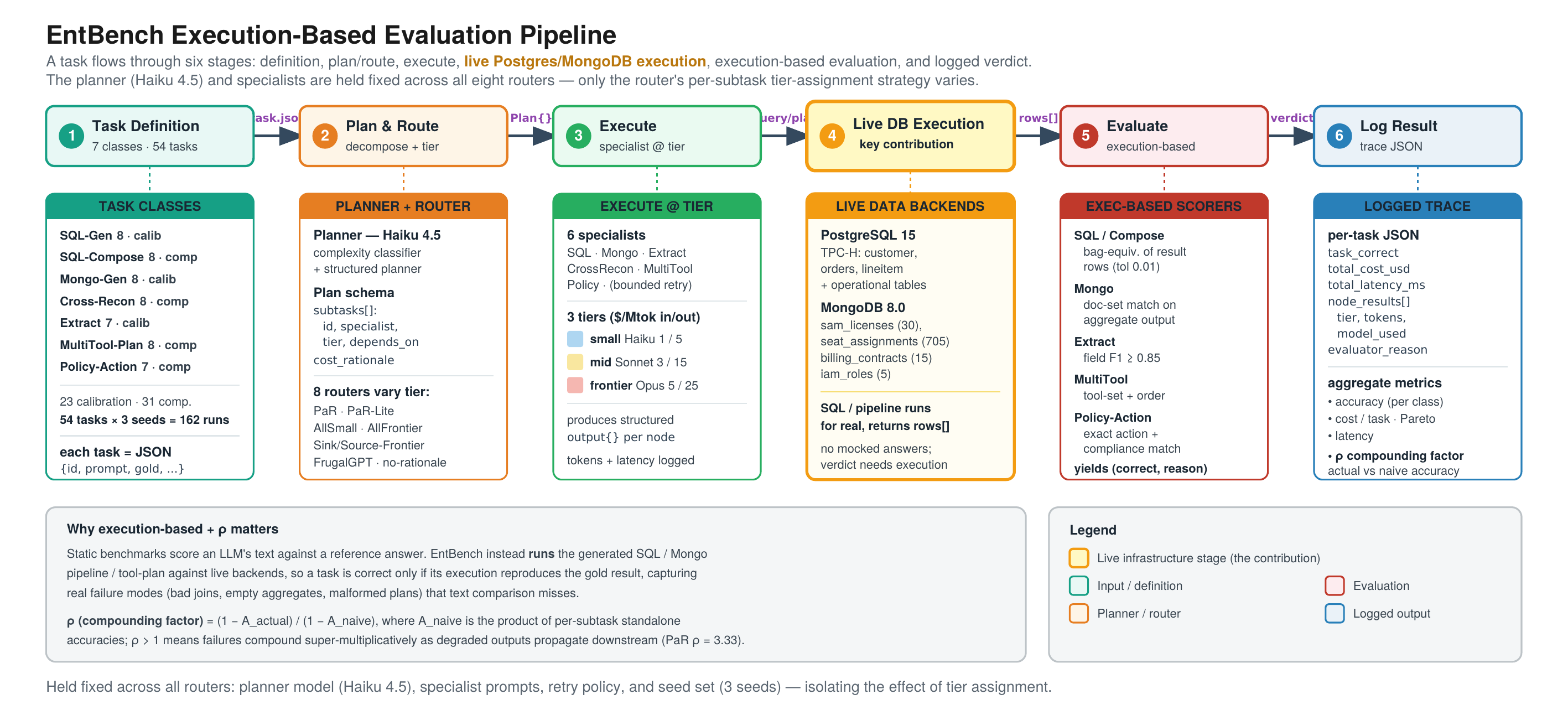}
\caption{EntBench execution-based evaluation pipeline. A task flows through six
stages, executing generated SQL/MongoDB against live backends so a task is
correct only if its execution reproduces the gold result---capturing failure
modes (bad joins, empty aggregates, malformed plans) that text comparison misses.
The planner, specialist prompts, retry policy, and 3 seeds are held fixed across
all eight routers; only per-subtask tier assignment varies.}
\label{fig:pipeline}
\end{figure*}

Grading runs the work rather than scoring text. Generated SQL executes against a
live PostgreSQL~15 database (TPC-H plus operational tables), compared by bag
equivalence (multiset match, ignoring row order) and correct only if the returned
rows match the gold result within tolerance; MongoDB pipelines run against a live
MongoDB~8.0 instance, matched on aggregate output; extraction is scored by
field-level F1, multi-tool plans by tool set and ordering, and policy actions by
exact action plus compliance match. Fig.~\ref{fig:pipeline} shows the pipeline.
Execution catches bad joins, empty aggregates, and malformed plans that text
comparison would accept as correct.

\begin{table}[t]
\centering
\caption{EntBench Task Classes (54 Tasks, Seven Classes)}
\label{tab:classes}
\begin{tabular}{@{}llll@{}}
\toprule
Class & Tasks & Type & Evaluator\\
\midrule
SQL-Gen        & 8 & Capability     & Execution bag equiv.\\
Mongo-Gen      & 8 & Capability     & Execution bag equiv.\\
Extract        & 7 & Capability     & Field-level F1\\
SQL-Compose    & 8 & Compositional  & SQL + consumer check\\
Cross-Recon    & 8 & Compositional  & Multi-source, 3-stage\\
MultiTool-Plan & 8 & Compositional  & Tool set + ordering\\
Policy-Action  & 7 & Compositional  & Exact match + vocab\\
\midrule
Total          & 54 & &\\
\bottomrule
\end{tabular}
\end{table}

\section{Experimental Setup}\label{sec:setup}
We compare eight routers, listed in Table~\ref{tab:routers}, holding the planner
and all specialist prompts fixed so only the tier-assignment strategy varies. All
monetary amounts are reported in United States dollars (USD), and all per-task
cost is in USD per task; accuracy is reported in percent and accuracy gaps in
percentage points (pp). The three tiers map to Claude Haiku~4.5 (small, USD 1/5
per million input/output tokens), Claude Sonnet~4.6 (mid, USD 3/15), and Claude
Opus~4.7 (frontier, USD 5/25). Every router runs all 54 tasks across 3 random
seeds, so each of the seven primary routers contributes 162 runs
($7\times162=1{,}134$ planned, less one failed SourceFrontier run that was
dropped), and the PaR-no-rationale ablation adds the eight Cross-Recon tasks
across three seeds (24), for 1,157 evaluations in total. PaR's measured plan mix
is 45\% small, 46\% mid, and 8\% frontier, averaging 1.8 subtasks per plan.

For reproducibility we pin exact model versions---small
\texttt{claude-haiku-4-5-20251001}, mid \texttt{claude-sonnet-4-6}, frontier
\texttt{claude-opus-4-7}---and run specialists and planner at the API default
temperature. Because the API is non-deterministic, each of the three seeds
denotes an independent repetition rather than a fixed RNG state; we report means
across seeds with task-level bootstrap confidence intervals (95\% CIs via paired
bootstrap, 10,000 resamples, task as the resampling unit, seeds averaged). All
planner and specialist prompts, the tool schema, and the grading harness are in
the released repository.

\begin{table}[t]
\centering
\caption{The Eight Routing Strategies Evaluated on EntBench}
\label{tab:routers}
\begin{tabular}{@{}ll@{}}
\toprule
Router & Description\\
\midrule
PaR           & Proposed: planner assigns a tier per subtask at plan time\\
PaR-Lite      & Cascaded: Haiku classifier skips planner on simple tasks\\
AllFrontier   & Every node on Opus 4.7 (accuracy upper bound)\\
AllSmall      & Every node on Haiku 4.5 (cost lower bound)\\
SinkFrontier  & Frontier on terminal nodes only, small elsewhere\\
SourceFrontier& Frontier on root nodes only, small elsewhere\\
FrugalGPT     & Small first; scorer escalates low-confidence outputs\\
PaR-no-rat.   & Ablation: PaR without the cost-rationale field\\
\bottomrule
\end{tabular}
\end{table}

\section{Design Rationale}\label{sec:rationale}
\textit{Execution-based grading over a rubric.} An LLM-as-judge rubric is
cheaper, but it injects a grader model whose own mistakes contaminate the routing
signal. We grade by execution instead: generated SQL and MongoDB pipelines run
against live databases, extractions scored by field-level F1, trading operational
complexity for an objective verdict that leans on no grader model.

\textit{Retry semantics as a fixed control.} Retries confound router
comparisons: retry harder and a router looks both more accurate and more
expensive of its tier choices. We fix one policy for every router: retry at the
assigned tier up to three times, escalate one tier once, then fail the node and
propagate a null. Accuracy and cost differences then reflect tier assignment, not
retry effort.

\textit{Prompt caching, evaluated and not adopted.} Caching and node-skipping we
evaluated but our per-specialist prompts fall below the provider's cacheable
minimum, so we do not rely on them.

\textit{Faithful baseline construction.} An unfaithful baseline silently flatters
the proposed method. Our first cascade accepted the first tier's output whenever
a confidence value was present, but the specialists emitted no usable confidence,
so it never escalated and collapsed onto small-tier behavior. We replaced it with
a faithful FrugalGPT-style cascade whose lightweight scorer assigns low-confidence
cases and escalates, charged at its own cost, more expensive and less accurate
than PaR, the honest comparison.

\textit{PaR-Lite as a deliberate control.} We expected PaR-Lite to keep most of
PaR's benefit at lower planning cost by skipping the planner onto a cheap
classifier where labels seem simple. It did not: it matches the cheapest routers
on accuracy but keeps PaR-level cost, which points to workflow-level planning, not
the classifier, as the source of the benefit.

\section{Results}\label{sec:results}
Table~\ref{tab:results} reports accuracy and cost for every router on the tasks
that carry a routing signal, which is the full set excluding SQL-Compose. We
exclude that class because it scores 0\% for all routers: 79\% of its failures
happen at the first SQL-generation step, a specialist-quality bottleneck that no
tier assignment can fix. We kept it in the benchmark because a uniform failure
that routing cannot touch is itself informative, but it does not belong in the
routing analysis.

\begin{figure}[t]
\centering
\includegraphics[width=\columnwidth]{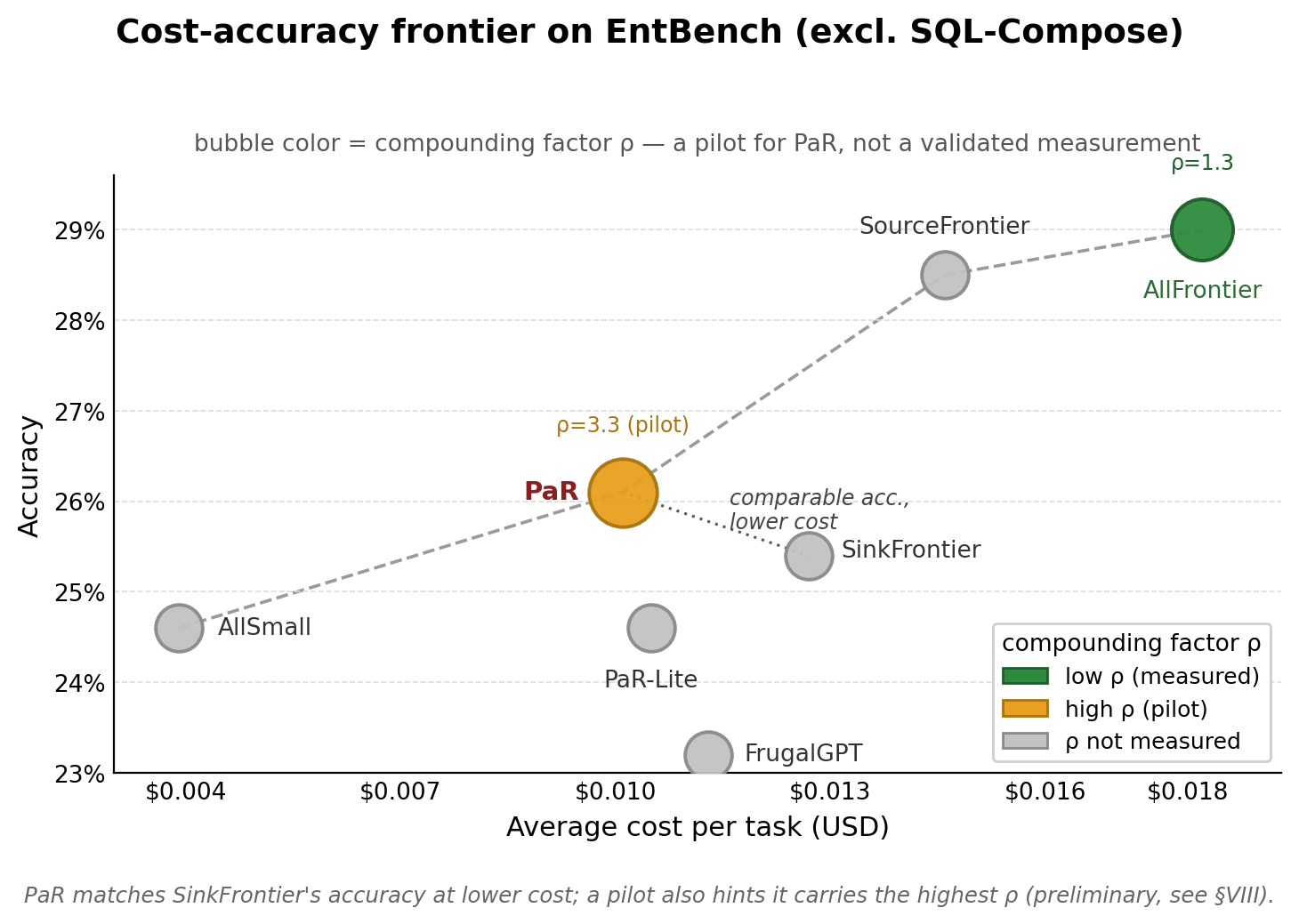}
\caption{Cost--accuracy frontier on EntBench (excluding SQL-Compose). Bubble color
encodes the compounding factor $\rho$ (a pilot value for PaR, measured for
AllFrontier, unmeasured otherwise). PaR sits on the Pareto frontier at comparable
cost to the sink-frontier heuristic.}
\label{fig:frontier}
\end{figure}

Paired accuracy gaps are not statistically separable here: the gap to AllFrontier
is $+2.5$~pp (CI $[-4.3,+9.9]$) and to SourceFrontier $+1.9$~pp, both spanning
zero. The benchmark separates routers on cost, not accuracy, and that is our
claim. PaR is significantly cheaper than AllFrontier ($-0.0074$~USD, CI
$[-0.0101,-0.0049]$) and SourceFrontier ($-0.0036$~USD, CI $[-0.0054,-0.0015]$)
at no significant accuracy cost, on both task sets. Relative to SinkFrontier, PaR
reaches comparable accuracy (26.1\% versus 25.4\%) at comparable cost. Under
Eq.~(1) at $\alpha=1$, PaR delivers about 1.6 times the accuracy-per-dollar of
AllFrontier and 1.3 times that of SinkFrontier, and ranks first among the four
accuracy-competitive routers at every $\alpha$ tested. Fig.~\ref{fig:frontier}
plots the full frontier.

\begin{table}[t]
\centering
\caption{Per-Router Results Excluding SQL-Compose ($n=138$ per Router, Three Seeds)}
\label{tab:results}
\begin{tabular}{@{}lrrr@{}}
\toprule
Router & Acc. & Avg Cost (USD) & vs.\ PaR\\
\midrule
AllFrontier    & 29.0\% & 0.0182 & $+2.9$pp, $+80\%$\\
SourceFrontier & 28.5\% & 0.0146 & $+2.4$pp, $+45\%$\\
PaR            & 26.1\% & 0.0101 & baseline\\
SinkFrontier   & 25.4\% & 0.0127 & $-0.7$pp, $+26\%$\\
AllSmall       & 24.6\% & 0.0039 & $-1.5$pp, $-61\%$\\
PaR-Lite       & 24.6\% & 0.0105 & $-1.5$pp, $+4\%$\\
FrugalGPT      & 23.2\% & 0.0113 & $-2.9$pp, $+12\%$\\
\bottomrule
\end{tabular}
\end{table}

Table~\ref{tab:perclass} gives the per-class breakdown for the primary routers.
On SQL-Gen every router converges near 50\% and tier choice barely matters. On
Mongo-Gen the frontier-heavy routers pull ahead modestly. PaR matches the
frontier-heavy routers on Extract, Policy-Action, and Cross-Recon while routing
much of those subtasks to cheaper tiers, the source of its cost advantage. That
ranking holds across cost weightings: PaR is first in utility (1) at every
$\alpha \in \{0.5, 1, 1.5, 2\}$.

\begin{table}[t]
\centering
\caption{Per-Class Accuracy (\%) by Router, Averaged Over Three Seeds}
\label{tab:perclass}
\begin{tabular}{@{}lrrrrrr@{}}
\toprule
Class & PaR & AllF & Src & Sink & AllS & Frug\\
\midrule
SQL-Gen        & 50.0 & 50.0 & 58.3 & 50.0 & 50.0 & 50.0\\
Mongo-Gen      & 25.0 & 37.5 & 39.1 & 37.5 & 25.0 & 25.0\\
Extract        & 28.6 & 28.6 & 28.6 & 28.6 & 23.8 & 19.0\\
Policy-Action  & 28.6 & 28.6 & 28.6 & 23.8 & 28.6 & 28.6\\
Cross-Recon    & 12.5 & 12.5 & 12.5 & 8.3  & 12.5 & 12.5\\
MultiTool-Plan & 12.5 & 16.7 & 4.2  & 4.2  & 4.2  & 12.5\\
SQL-Compose    & 0.0  & 0.0  & 0.0  & 0.0  & 0.0  & 0.0\\
\bottomrule
\end{tabular}
\end{table}

\subsection{Planner Behavior in Practice}
Table~\ref{tab:planbehavior} reports how the planner actually assigns tiers,
measured across the 162 PaR runs. The complexity classifier labels 43\% of
task-runs \textsc{simple} (a single small-tier subtask, with no full planner
call) and 57\% \textsc{complex}; overall the planner emits 1.8 subtasks per plan
and a tier mix of 45\% small, 46\% mid, and 8\% frontier. The per-class breakdown
shows the planner reasoning about difficulty rather than defaulting to one tier:
single-step capability classes (SQL-Gen, Mongo-Gen, Extract) stay almost entirely
on the small tier, while compositional classes escalate---SQL-Compose and
Cross-Recon route the majority of subtasks to mid, and only the genuinely hard
reconciliation and multi-tool subtasks reach frontier. This is the mechanism
behind PaR's cost position: it spends the frontier tier where the task structure
warrants it and nowhere else. It also disambiguates the classifier from the
planner---the 43\% simple short-circuit accounts for the cheap single-step
classes, while the tier reasoning on the 57\% complex tasks is what places PaR on
the frontier.

\begin{table}[t]
\centering
\caption{Planner Tier Assignment by Task Class (PaR, 162 Runs). Tier Columns Give the Percentage of Subtasks Routed to Each Tier.}
\label{tab:planbehavior}
\begin{tabular}{@{}lrrrr@{}}
\toprule
Class & Avg subtasks & Small \% & Mid \% & Frontier \%\\
\midrule
SQL-Gen        & 1.0 & 75 & 25 & 0\\
Mongo-Gen      & 1.0 & 88 & 12 & 0\\
Extract        & 1.0 & 95 & 5  & 0\\
Policy-Action  & 1.2 & 40 & 60 & 0\\
SQL-Compose    & 2.1 & 12 & 78 & 10\\
Cross-Recon    & 2.8 & 25 & 55 & 19\\
MultiTool-Plan & 3.3 & 50 & 42 & 8\\
\bottomrule
\end{tabular}
\end{table}

\section{A Preliminary Look at Compounding}\label{sec:compounding}
Compounding is a property of any routing strategy, not peculiar to PaR. This
section is a preliminary probe, not a validated measurement: the numbers come
from a small pilot, cover two of eight routers, and carry a confound we make
explicit. We define a compounding-error factor as the ratio of a router's actual
end-to-end failure rate to what a naive independence assumption predicts; above
one means failures correlate along the dependency chain and amplify. For a
workflow of $k$ nodes with standalone per-node accuracies $a_i$ (from the
capability-class sweep),
\begin{equation}
\rho \;=\; \frac{1-A_{\mathrm{actual}}}{1-A_{\mathrm{naive}}},
\qquad
A_{\mathrm{naive}} = \prod_{i=1}^{k} a_i,
\end{equation}
where $A_{\mathrm{actual}}$ is measured end-to-end accuracy and
$A_{\mathrm{naive}}$ the accuracy predicted under independent node failures; the
reported $\rho$ averages these per task over the pilot set.
Table~\ref{tab:rho} reports the two routers that bracket the spectrum of
cheap-tier mixing, using a 21-task capability-class pilot as the standalone-accuracy
baseline.

Two caveats keep this preliminary. First, $A_{\mathrm{naive}}$ is built from
capability-task standalone accuracies but used to predict compositional-task
end-to-end accuracy; if the embedded subtasks are harder than the standalone
ones, $A_{\mathrm{naive}}$ is inflated and part of the apparent compounding
reflects that mismatch, not propagation alone. Second, the two rows of
Table~\ref{tab:rho} are not on a common task population, so the contrast between
$\rho=3.33$ and $\rho=1.27$ is illustrative, not a measured spectrum. A full
54-task sweep across all 18 specialist-tier combinations on one shared task set
would turn this signal into a result; we leave it to future work
(Section~\ref{sec:future}).

\begin{table}[t]
\centering
\caption{Compounding Error ($\rho$). (*) $\rho$ Values From a 21-Task Pilot Sweep}
\label{tab:rho}
\begin{tabular}{@{}lrrrl@{}}
\toprule
Router & $\rho$ & Actual & Naive Pred. & Status\\
\midrule
PaR         & 3.33 & 25.0\%* & 77.5\%* & Pilot, high\\
AllFrontier & 1.27 & 27.8\%  & 43.1\%  & Measured, low\\
\bottomrule
\end{tabular}
\end{table}

\begin{figure}[t]
\centering
\includegraphics[width=\columnwidth]{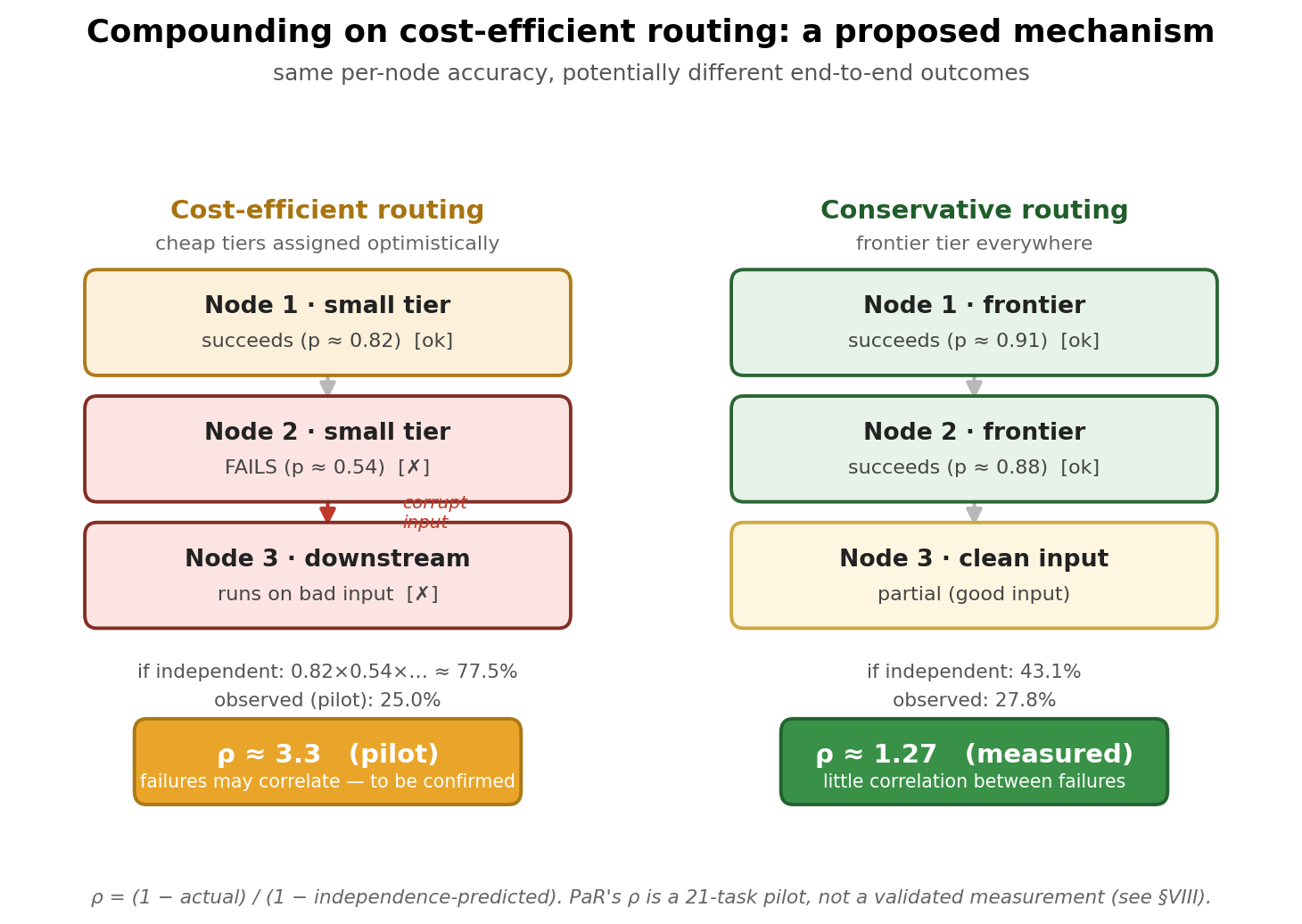}
\caption{Compounding-error mechanism (illustrative). Cheap-tier assignments (left)
can suffer cascade failures when a small-tier node fails and corrupts downstream
input; all-frontier routing (right) keeps per-node failure rare, so compounding
stays mild. PaR's $\rho$ is a 21-task pilot, not a validated measurement.}
\label{fig:compounding}
\end{figure}

\subsection{Ablation: What the Design Choices Buy}\label{sec:ablation}
Two ablations test whether PaR's components are load-bearing. Removing the
cost-rationale field, so the planner assigns tiers but no longer justifies them,
collapses the Cross-Recon class (24 task-seed evaluations) to 0\% accuracy at more
than four times PaR's cost; a targeted probe on one compositional class, but large
and consistent across seeds, which says the rationale field steers tier assignment
rather than being cosmetic. The second ablation, PaR-Lite, is a clean negative: it
matches the cheapest routers on accuracy (24.6\%) at PaR-level cost (USD 0.0105),
so the benefit comes from integrated planner reasoning over the whole workflow,
not a bolt-on classifier.

\section{Discussion and Threats to Validity}\label{sec:discussion}
The claim is modest but useful: moving tier selection into planning lets a
workflow reach a good point on the cost--accuracy frontier with no trained router
and no labeled data, the benefit coming from the planner's holistic view rather
than any single heuristic. PaR is therefore not competing with learned routers on
their terms; it is a lighter-weight route to a comparable result that drops into a
graph-based agent framework as a prompt change plus a dispatcher.

Several limitations bound these conclusions. The benchmark is 54 tasks, so
confidence intervals are wide ($\pm6$ points) and we lean on frontier position
rather than separable accuracy gains; absolute accuracies are low (22--29\%),
reflecting how hard execution-based grading is on compositional tasks. PaR's
compounding factor is a 21-task pilot, and the planner is one provider's model
assigning that provider's tiers, so a mixed-provider setting is untested.
Our baselines are heuristic and cascade routers. We do not compare against a
trained query- or agent-level router such as RouteLLM~\cite{routellm} or
MasRouter~\cite{masrouter}: those require training data that EntBench does not
provide, and a faithful port to a per-subtask, execution-graded setting is its own
undertaking. We therefore frame PaR as a training-free alternative to learned
routers, not a head-to-head winner over them. All limitations are disclosed rather
than smoothed over---including that the cost separation reverses against AllSmall:
PaR is significantly more expensive than AllSmall ($+0.0073$~USD, CI
$[+0.0049,+0.0098]$) at equal accuracy, so its contribution is the routing
mechanism at frontier-matched accuracy, not a universal cost win.

\section{Future Work}\label{sec:future}
Several directions follow. The most immediate is scale: growing EntBench toward
its full 300-task design would tighten the intervals enough to separate routers
now within noise, concentrating new tasks in the routing-sensitive
middle-difficulty classes. A second is completing the compounding measurement: a
full standalone sweep across all 18 specialist-tier combinations on one shared
task set would replace the pilot $\rho$ values with grounded numbers for every
router. Third is uncertainty-aware tier assignment: a planner that flags a subtask
as both error-prone and high-fan-in could assign a stronger tier precisely because
its failure would cascade. A mixed-provider setting---one provider's planner
assigning another's tiers---is untested and is the most interesting confound to
resolve.

\section{Conclusion}
We presented Planner-as-Router, a training-free pattern that folds model-tier
selection into planning, assigning a tier to every subtask jointly across the
dependency DAG before any specialist runs. On EntBench---54 enterprise tasks
graded against live databases---PaR stays on the cost--accuracy frontier: it
matches a sink-frontier heuristic at comparable cost and a faithful FrugalGPT
cascade at lower cost, and cuts cost 44\% against all-frontier routing for a
2.9-point accuracy drop. We also report a preliminary compounding-error signal
that cheap-tier mixing may carry a propagation penalty flat reporting would hide,
offered as a hypothesis for future measurement. PaR, EntBench, and all code are
open source at \url{https://github.com/vsingh45/par-entbench}.

\section*{Acknowledgment}
The authors used a generative AI system (Anthropic Claude) for language editing,
formatting, and coding assistance during manuscript preparation. All research
contributions, methods, data, and results are the authors' own work.

\end{document}